%% file: 0_main.tex
\documentclass{article}

\usepackage{arxiv}

\usepackage[utf8]{inputenc} 
\usepackage[T1]{fontenc}    
\usepackage{hyperref}       
\usepackage{url}            
\usepackage{booktabs}       
\usepackage{amsfonts}       
\usepackage{nicefrac}       
\usepackage{microtype}      
\usepackage{lipsum}		
\usepackage{graphicx}
\usepackage[square,numbers,sort&compress]{natbib}
\usepackage{doi}
\usepackage{color}
\usepackage{enumitem}
\usepackage{subcaption}

\setlist[itemize]{leftmargin=6pt, labelindent=0pt, itemindent=0pt}

\title{Delivering Science as a Service: Sci-Orchestra's Cloud-Native Approach to HPC}

\author{ 
    \href{https://orcid.org/0000-0001-8018-0547}{\includegraphics[scale=0.06]{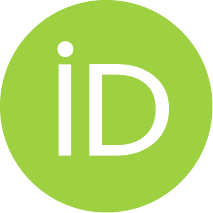}\hspace{1mm} Harinarayan Krishnan} \\
    Applied Math \& Computational Research Division \\
	Lawrence Berkeley National Laboratory \\
	One Cyclotron Road, Berkeley, CA 94720 \\
	\texttt{hkrishnan@lbl.gov} \\
	\And
    \href{https://orcid.org/0009-0003-7173-0174}{\includegraphics[scale=0.06]{orcid.pdf}\hspace{1mm}Jeffrey Donatelli} \\
    Applied Math \& Computational Research Division \\
	Lawrence Berkeley National Laboratory \\
	One Cyclotron Road, Berkeley, CA 94720 \\
	\texttt{jjdonatelli@lbl.gov} \\
    \And
    \href{https://orcid.org/0000-0002-9093-539X}{\includegraphics[scale=0.06]{orcid.pdf}\hspace{1mm}Shubhabrata Mukherjee} \\
    Applied Math \& Computational Research Division \\
	Lawrence Berkeley National Laboratory \\
	One Cyclotron Road, Berkeley, CA 94720 \\
	\texttt{smukherjee@lbl.gov} \\
	\And
	\href{https://orcid.org/0000-0002-7363-9468}{\includegraphics[scale=0.06]{orcid.pdf}\hspace{1mm}Daniela Ushizima} \\
    Applied Math \& Computational Research Division \\
	Lawrence Berkeley National Laboratory, \\
	Bakar Comp. Health Sciences Institute, UC San Francisco, \\
    Berkeley Institute for Data Science, UC Berkeley \\
	\texttt{*dushizima@lbl.gov} \\
}

\renewcommand{\headeright}{Technical Report}
\renewcommand{\undertitle}{Technical Report}
\renewcommand{\shorttitle}{\textit{Sci-orchestra}}

\hypersetup{
pdftitle={A template for the arxiv style},
pdfsubject={q-bio.NC, q-bio.QM},
pdfauthor={David S.~Hippocampus, Elias D.~Striatum},
pdfkeywords={First keyword, Second keyword, More},
}

\begin{document}
\maketitle

\begin{abstract}
The increasing complexity of modern computational environments often burdens researchers with infrastructure management, authentication protocols, and container deployments. We present Sci-Orchestra, a layered orchestration framework designed to fully automate experimental workflows, allowing scientists to prioritize scientific discovery over backend operations. By abstracting execution through an API-driven interface, the system assumes responsibility for secure authentication, resource management, and scalable deployment across diverse high-performance computing environments using Kubernetes architectures. A key innovation of Sci-Orchestra is its autonomous marketplace, which serves as a catalyst for cross-institutional collaboration. Through an intuitive user interface, researchers can rapidly deploy and share specialized services via simple selections, eliminating the need for complex installations and technical setups. This modular infrastructure is specifically designed to facilitate industry partnerships as it provides a secure execution environment and allows external collaborators to test and validate proprietary tools without the need for source-code exchange. This ``black-box'' interoperability protects intellectual property while enabling seamless integration into broader scientific pipelines, ultimately accelerating the transition from laboratory prototypes to industrial-scale applications.
\end{abstract}

\keywords{Science Platform \and Orchestration \and Distributed Ecosystem \and Resource Management \and Kubernetes \and MCP }

\input{1_newcontent}

\bibliographystyle{unsrtnat}
\bibliography{2_references}  






\end{document}

%% file: 1_newcontent.tex
\section{Introduction}
As scientific discovery becomes fundamentally reliant on advanced computational models~\citep{krishnan2020towardsstack,ushizima_pycbir}, the heavy technical burden of deploying these environments has become a critical bottleneck, routinely forcing researchers to act as system administrators rather than scientists. To circumvent this problem, we introduce Sci-Orchestra: a highly automated, layered infrastructure framework engineered to seamlessly manage the entire lifecycle of experimental workflows. Anchored by a rigorous API-first design philosophy, Sci-Orchestra completely abstracts local computational overhead by providing a modular ecosystem for registering data, compute, and network functionalities. This paradigm shifts the burden of resource allocation from the user to the system, empowering researchers to provision, scale, and launch deeply complex scientific workflows instantly through a single, streamlined API call.

Operating as an advanced orchestration engine, Sci-Orchestra securely handles user authentication, container deployment, hardware resource mounting, and seamless interoperability between distinct analytical programs. By wrapping applications into web-accessible endpoints, the system decouples complex software dependencies from direct local installation. This architecture provides a powerful preview mechanism, allowing researchers to experience and evaluate tool features directly through common clients such as \textbf{web-browsers}, \textbf{command line tools} and \textbf{chatbots}. Furthermore, an automation-first strategy ensures that Sci-Orchestra can span the computational spectrum. On one end, it significantly lowers the barrier for deploying manual, one-off scientific applications; on the other, it provides the essential infrastructure required for autonomous agents to execute large, cross-facility workflows with minimal human intervention.

\begin{figure}[t!]
    \centering
    \includegraphics[width=1\linewidth]{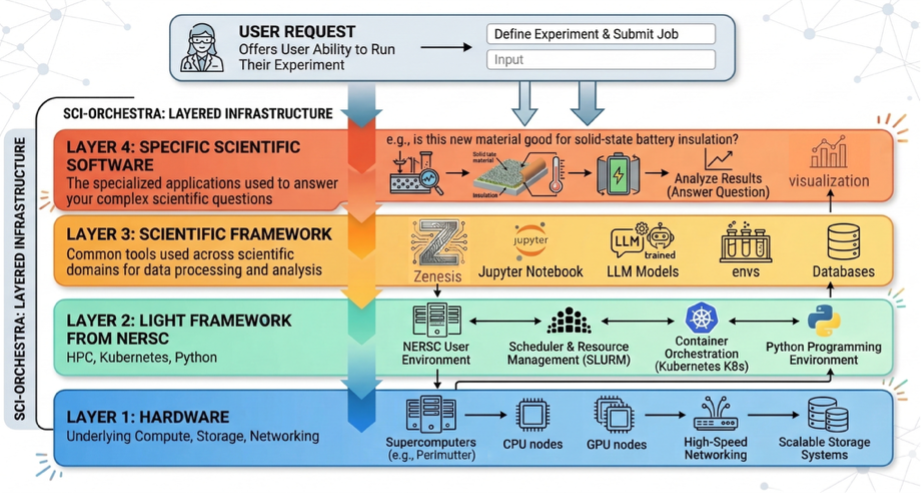}
    \caption{Sci-orchestra layered architecture and user interaction model enable seamless hardware and software integration delivered as services through a standard web-browser.}
    \label{fig:sciorchestra}
\end{figure}

As Figure~\ref{fig:sciorchestra} illustrates, the framework removes common pain points associated with deploying scientific technology stacks by managing the underlying infrastructure on behalf of the end user. This is achieved through two key insights: strictly separating the user experience from developer responsibilities, and employing a modular, service-discovery-first approach. These design choices maximize flexibility for developers managing computational resources while minimizing the learning curve for scientists utilizing the platform.

The contributions of Sci-Orchestra are particularly transformative for the integration of artificial intelligence and automated experimentation. By providing a robust, programmatic foundation, the framework allows machine learning models, large language models, and autonomous software agents to dynamically request High-Performance Computing (HPC) resources, orchestrate analytical pipelines, and iterate on data streams without traditional human-in-the-loop bottlenecks. This high-speed, programmatic orchestration is uniquely suited for dynamic, time-sensitive applications, such as autonomous materials discovery and the real-time analysis of live sensor data during active experiments. In these rapid-response domains, where immediate computational feedback and secure data handling are paramount, Sci-Orchestra empowers distributed research teams and autonomous AI frameworks to instantly provision isolated analysis environments. This enables the rapid modeling of dynamic physical phenomena and the real-time processing of complex scientific datasets within a fully reproducible, tightly controlled computational ecosystem.

\section{Background and Related Work}


Projects utilizing HPC platforms at U.S. national laboratories—such as the National Energy Research Scientific Computing Center (NERSC), the Argonne Leadership Computing Facility (ALCF), and the Oak Ridge Leadership Computing Facility (OLCF)—often possess niche computational requirements tailored to highly specific  projects. Scientific diversity dictates these bespoke configurations.

While a large portion of the research conducted at these institutions is unclassified and focused on fundamental open science, they simultaneously maintain significant national security portfolios. To accommodate this dual mission, these facilities must enforce overarching, strict operational and network security policies capable of supporting sensitive workloads, including Controlled Unclassified Information (CUI) and secure data enclaves. As a result, developers are routinely required to execute workflows within the pre-approved orchestration systems and software stacks established by the computing centers. While these tightly governed environments are highly optimized for traditional, large-scale simulation workloads, the rapid integration of artificial intelligence, machine learning, and Internet of Things (IoT) edge devices introduces dynamic, real-time architectural requirements that stretch the limits of the current paradigms.

These modern scientific applications also range from high-speed streaming cameras, such as those at the Advanced Light Source (ALS)\citep{hoying:2023} that require real-time data routing, to interactive machine learning engines, such as tsuchinoko\citep{lblcamera_tsuchinoko} and pyCBIR\citep{ushizima_pycbir,pycbir:2018}. Furthermore, complex deep learning workflows—including Rhizonet\citep{rhizonet} and Zenesis\citep{zenesis2025} for scientific image analysis, or BERTeley\citep{Berteley} for natural language processing. Several tools rely heavily on persistent web accessibility, dynamic resource provisioning, and near real-time interactivity. Accommodating these workloads exposes structural limitations in the conventional HPC software stack and requires the development of flexible orchestration solutions that can operate securely within these restricted environments.

The challenge of automating such scientific workflows has driven the development of numerous open-source orchestration projects. Traditional scientific \textbf{workflow managers}, such as Nextflow~\citep{nextflow2017}, Snakemake~\citep{snakemake2012}, and Pegasus~\citep{pegasus2015}, are robust systems designed primarily to execute complex, static data processing pipelines. These platforms possess strong support for HPC backends, seamlessly translating tasks into batch jobs submitted to workload managers like Slurm or PBS. However, because these tools are fundamentally optimized for non-interactive, batch-oriented execution, they require researchers to define infrastructure parameters via static code. Consequently, they do not provide the continuous service-hosting capabilities required to deploy live, interactive dashboards or API-driven web applications directly to the end user.

In the realm of interactive, \textbf{browser-based tools}, platforms such as JupyterHub~\citep{jupyter2016} and BinderHub~\citep{binder2018} have established a strong precedent. These systems natively utilize Kubernetes to orchestrate containerized environments, allowing users to access data science notebooks~\citep{holdgraf2017portable,MRS:2020:concrete} directly from their browsers. While primarily cloud-native, the community has developed extensions such as BatchSpawner to allow JupyterHub to interface with traditional HPC environments. Despite this capability, these platforms are generally highly specialized for launching isolated, single-user notebook servers. They are not designed to natively orchestrate multi-component service architectures, deploy custom sidecar container applications, or manage persistent, event-driven research lifecycles that span multiple distinct computational facilities.

Other general-purpose \textbf{orchestrators}, such as Apache Airflow~\citep{airflow2021}, offer robust event management and complex service scheduling but are heavily biased toward enterprise cloud environments. Adapting these cloud-first tools to the strict security policies, constrained networking, and rigid resource allocation rules of security-constrained HPC centers poses a significant engineering challenge. Consequently, a critical gap remains: the scientific community lacks a platform that combines the interactive, browser-first user experience of a cloud framework with the ability to dynamically deploy complex, interconnected microservices into highly restrictive hybrid and HPC environments.

Addressing these compounding gaps is only now achievable due to the maturation of three key technological enablers: Containers, Infrastructure as Code (IaC) driven Deployment/Serverless Platforms, and AI/ML efforts, as described below.

\paragraph{Containers} Over the past decade, the advent of containerization~\cite{boettiger2015introduction,merkel2014docker} has successfully decoupled the software ecosystem from the low-level toolchains specific to individual compute facilities. This paradigm shift enables developers to deploy applications across multiple heterogeneous environments using a single, unified build, eliminating the need to compile custom software for every unique system. Consequently, resource providers—whether HPC centers, commercial clouds, edge networks, or Scientific User Facilities (SUFs)—can focus entirely on provisioning the underlying hardware platform and device drivers, relying on the containerized software stack to manage its own dependencies. While major cloud providers readily integrated container orchestration into their core offerings, the adoption of these technologies within HPC environments was initially impeded by the severe security vulnerabilities associated with running containers that required root privileges. However, the development of runtimes such as Singularity/Apptainer and Podman successfully mitigated these concerns by leveraging user namespaces, isolating code execution without compromising system-wide security. With a growing number of HPC centers now natively supporting user-namespace container technologies, Sci-Orchestra is uniquely positioned to deploy stable, reproducible workflows that remain entirely independent of specific computational facility configurations.

\paragraph{Infrastructure as Code} Infrastructure as Code (IaC) is the programmatic methodology of managing and provisioning computing resources through machine-readable, declarative definition files, rather than relying on physical hardware setup or interactive configuration interfaces~\cite{morris2020infrastructure}. By treating infrastructure as version-controllable source code, IaC ensures that complex deployment environments remain strictly reproducible, scalable, and automated. This paradigm shifts resource management from manual system administration to standardized API execution, enabling orchestration platforms to dynamically assemble network, storage, and compute assets strictly on demand. 

In fact, cloud environments have evolved to enable creation of two mechanisms for requesting resources~\citep{farias2023iac}. The first is a serverless approach where clients have minimal requirements and leave the rest of resource provisioning to the resource providers. The second is a more fine-grained solution that adds a resource provisioning API. Solutions such as AWS Cloud Development Kit (CDK) and Terraform/OpenTofu enable users to specifically request resources to target deployment systems. Normally, HPC systems lack the granularity of their cloud-based counterparts, until recently when Kubernetes was added as an additional deployment target. In part, this community is recognizing that workflows and tools themselves require an ecosystem that oversees progress and controls decision making. Now, HPC workloads can be paired with databases that track state and progress, dashboards that provide visual feedback, and a plethora of tools for better scientific analytics, bringing them closer in parity to  modern cloud-like architecture models. Sci-Orchestra, through its service-discovery-based design, enables users to take advantage of deploying services, workflows, and all necessary applications using an API-first approach, moving towards an HPC-IaC model.
\paragraph{ML/AI} Finally, this infrastructure sets up the landscape for AI agents to take automation a step further. This includes the ability to make one-to-many science plans and authorize them through an industry-standard security model—such as OAuth 2.0 Dynamic Client Registration~\citep{rfc7591} with SPIFFE/SPIRE support~\citep{posta_spiffe}—to accommodate an increasingly agent-driven environment. Sci-Orchestra is built to power Model-Context Protocol (MCP)~\citep{posta_spiffe} services, and also what that might potentially mean in an HPC-style environment where observability is paramount.   

While various industry-leading technologies support subsets of these capabilities, Sci-Orchestra integrates all three within a strict \textbf{service-discovery-first} architecture. In this paradigm, the orchestration engine dynamically identifies, registers, and binds available computational and data services at runtime, entirely bypassing the need for static infrastructure configurations or hard-coded network routes. Because scientific workloads face highly variable hardware availability and strict institutional security policies, this \textbf{dynamic routing} capability delivers the comprehensive flexibility researchers require to seamlessly operate within a rapidly shifting technological landscape.

\section{System Architecture and Event Management}
At its core, Sci-Orchestra operates on a flexible Kubernetes-based event management architecture. The prototype system is cloud-based, currently running in Kubernetes environments such as NERSC SPIN, and comprises three major components operating together: an application programming interface, a backend database, and a modular dedicated orchestration engine. User requests are stored in the database and processed by the main orchestration engine, which evaluates resource requirements and coordinates with the appropriate computational backends. This decoupled design ensures that the orchestrator remains independent of any single hosting platform and can be seamlessly adapted to varied environments such as cloud Kubernetes, the National Research Platform (NRP)~\cite{nrp_project, smarr2018pacific} -- a federally funded, distributed Kubernetes federation -- or any dedicated hosting environment. The motto for Sci-Orchestra is: if it has an API, then we can connect to it. 

\begin{figure}[b!]
    \centering
    \includegraphics[width=1\linewidth]{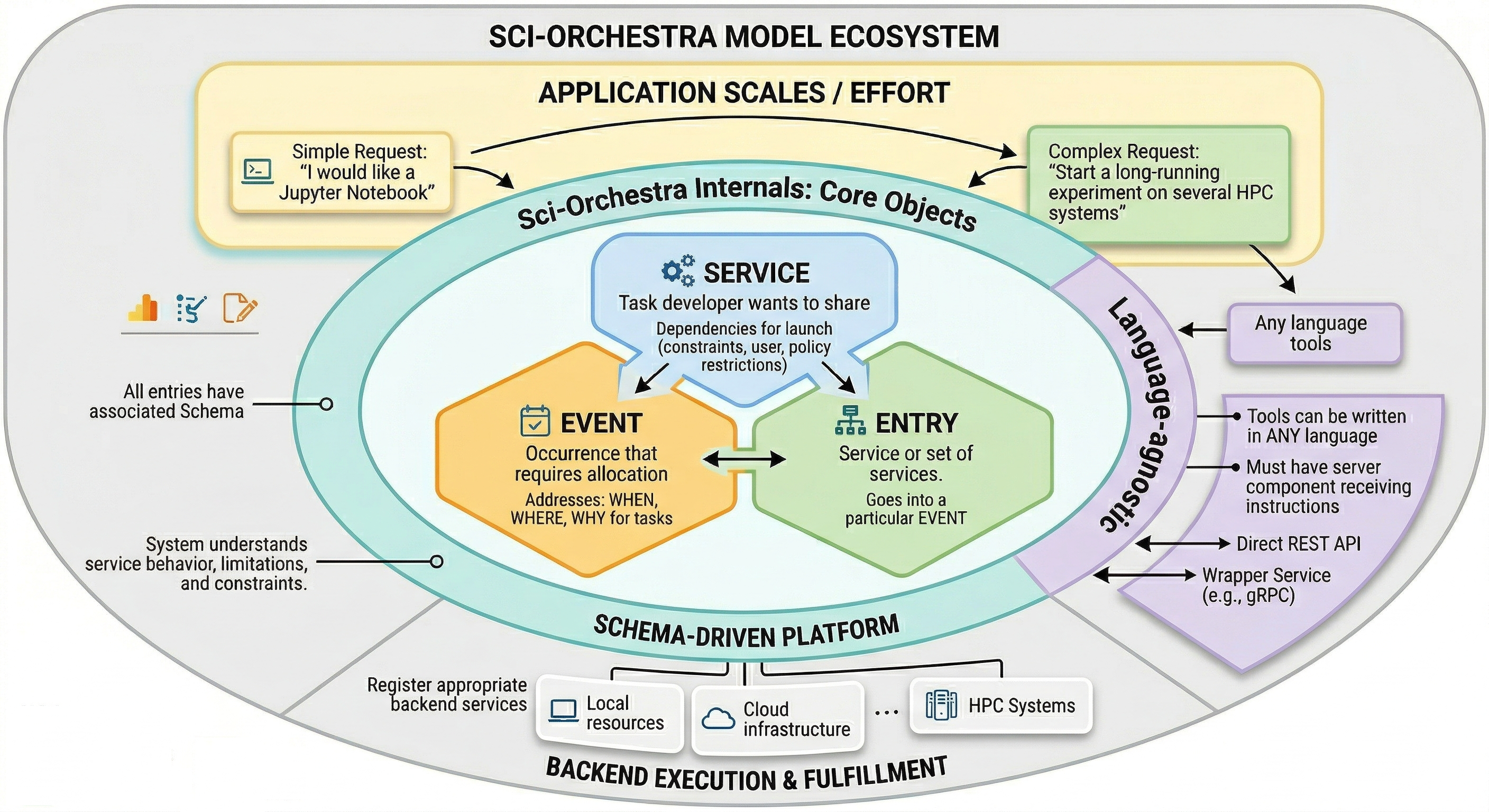}
    \caption{A detailed architectural diagram of the Sci-Orchestra Model Ecosystem, illustrating how user requests (simple to complex) are processed through core ``Schema-Driven'' and ``Language-agnostic'' internals (service, event, entry) to register and execute tasks on backend resources such as cloud or HPC systems.}
    \label{fig:sciorchestra-model}
\end{figure}

\subsection{Sci-Orchestra Model}
Disparate technologies and fragmented infrastructures traditionally pose significant challenges to deploying robust scientific workflows. Computational platforms across research institutions frequently operate as distinct systems, demanding interaction with low-level tools and the creation of custom deployment scripts. The recent adoption of standardized deployment systems, specifically Kubernetes, at major HPC facilities provides a tractable mechanism to control orchestration across these diverse environments.

Recognizing this shift, Sci-Orchestra operates as a Science-Platform-as-a-Service. The framework provides higher-level abstractions to uniquely route unique raw algorithms and workflows integrated with modern, cloud-native web applications. To support this, Sci-Orchestra delivers fine-grained user access control, centralized scheduling for distributed resources, and a programmatic API for remote execution. These features allow researchers to provision secure, production-grade applications and share them with collaborators while abstracting underlying infrastructure bottlenecks.

At its foundational layer, the architecture utilizes a schema-driven approach. Every component within Sci-Orchestra is associated with a defined schema, enabling the system to automatically parse service behaviors, required inputs, and operational constraints. The platform is intentionally language-agnostic. This contract based design enables Sci-Orchestra to communicate the necessary requirements to run and operate different workflows. 


The internal execution model of Sci-Orchestra relies on three core components, as illustrated in Fig. 2. The first component is the Service, which defines the specific task or application a developer intends to deploy, encompassing all necessary dependencies and policy constraints. The second component is the Event, an operational instance requiring a computational allocation that dictates the timing, location, and parameters of the executed tasks. The third component is the Entry, which serves as the concrete instantiation of one or more services injected into a specific Event.

To contextualize these abstractions, consider a researcher provisioning an interactive Jupyter Notebook environment. The Service functions as the static blueprint; it defines the base Jupyter container image, specifies required software libraries (such as PyTorch or TensorFlow), and establishes hardware constraints, such as a requirement for GPU acceleration. When the researcher initiates a session, the platform generates an Event. This Event bounds the operational context, establishing the execution timestamp, designating a specific computational backend (e.g., a NERSC Perlmutter compute node), and mapping the required distributed storage volumes. Finally, the Entry represents the live execution: the instantiated Kubernetes pods actively running the notebook and the securely routed, dynamic URL (e.g., https://jupyter-123.sciorchestra.org) that provides the user with secure, browser-based access to the environment.


These concepts correspond to varying scales of computational effort. A user request, ranging from a single Jupyter Notebook deployment to a persistent multi-node experiment across distributed facilities, is uniformly processed as an Event that registers the appropriate backend Services. Furthermore, the platform employs a service-discovery architecture. Instead of relying on hard-coded connections, the system dynamically expands at intersection points between compute and data resources. Decoupling the software application from the underlying orchestrator backend allows Sci-Orchestra to dynamically discover and provision the appropriate execution environment for diverse scientific tasks.

\subsection{Sci-Orchestra Ecosystem}
The core model itself is only a part of the equation for Sci-Orchestra. As illustrated in Table~\ref{tab:sci_orchestra_tech}, to transform it into a production-ready vehicle, Sci-Orchestra provides the ability to authenticate and authorize each event, and share events with collaborators through a security-first authentication process. Each service that has an external-facing entry is then provided a web-accessible entrypoint with an HTTPS certificate to ensure that the site is secure. This is all done as part of the deployment process and is transparent to the developer of a service. Additionally, authorization is event-based, and users are given access to resources based on the collection of entries in an event. 

The API-first deployment process simplifies user interaction. For example, a user can initiate a request via an API call (e.g., \texttt{/start/jupyter\_notebook}). The platform processes this request, provisions the resource on the backend, and the user can query the status until completion. Once ready, the API returns a secure, locked-down URL (e.g., \texttt{https://abc123.sciorchestra.org}) allowing immediate browser access. 

This model excels at service composition. Consider a combined scientific project requiring both a logging service (e.g. Weights \& Biases~\cite{biewald2020experiment}) and a computational interface (e.g. JupyterHub). A user can first send an API call to start the logging service, which returns a secure URL. A subsequent API call can then launch the .ipynb Notebook, passing the logging service's URL and credentials as a JSON payload. Because both components are launched within the same Sci-Orchestra event, they seamlessly share information and coexist under the same authentication and authorization umbrella.

A hierarchical service discovery model is employed to manage these deployments across diverse hardware. When the API schedules a request for an event containing a specific service (e.g., a Jupyter Notebook requiring GPU resources), the main orchestration engine evaluates the request. It acts as a router, delegating the workload to specialized sub-orchestrators based on resource requirements. For example, a user wants to bypass CPU-only resources to have their request delegated to an orchestrator connected to a GPU-capable system (such as NERSC Perlmutter), while ensuring web accessibility constraints are met. This design allows the main orchestrator to provide access to vast resources without needing to know the low-level deployment logic of every individual location. It also enables complex cross-system communication. If a Jupyter notebook on a compute node needs to access a database hosted on a separate cloud partition such as NERSC Spin, Sci-orchestra-centric developers can create custom service templates that automatically establish SSH tunnels, bridging systems that cannot communicate directly.

\begin{table*}[t]
    \centering
    \renewcommand{\arraystretch}{1.4}
    \caption{Key technologies that power the Sci-Orchestra framework, with layered structure illustrated in Fig.~\ref{fig:sciorchestra}}
    \label{tab:sci_orchestra_tech}
    \begin{tabular}{|p{3cm}|p{2.5cm}|p{5cm}|p{5cm}|}
        \hline
        \textbf{Technology / Component} & \textbf{Infrastructure Layer} & \textbf{Primary Role within Sci-Orchestra} & \textbf{Key Benefit to the Researcher} \\
        \hline
        \textbf{Kubernetes} & Layer 2: Light Framework & Automates the deployment, scaling, and operation of application containers using namespaces, sidecar configurations, service discovery, and load balancing. & Completely abstracts the backend, allowing users to launch complex, multi-container environments without ever touching a command-line interface. \\
        \hline
        \textbf{FastMCP} & Layer 3: Common Software & Acts as a universal, standardized protocol bridge connecting the core orchestration engine and artificial intelligence models to independent scientific applications. & Enables rapid, plug-and-play integration of new scientific tools into the ecosystem without requiring developers to write complex, custom API wrappers. \\
        \hline
        \textbf{Docker \& Container Caching} & Layer 2: Light Framework & Packages specific scientific software into isolated, standardized environments, while local host caching stores these images directly at computing centers such as NERSC. & Guarantees environmental consistency across different experiments and drastically reduces the time it takes to launch a tool from the browser. \\
        \hline
        \textbf{REST API \& Python} & Layer 2: Light Framework & Serves as the communication backbone for the orchestration worker, managing the backend database, user portfolios, and the event-driven lifecycle of experiments. & Translates simple web browser clicks into complex backend orchestration commands, making event and lifecycle management seamless. \\
        \hline
        \textbf{Federated Authentication} & Layer 3: Access & Handles secure identity verification using Google credentials or institutional logins, generating secure access to event-specific URLs. & Provides frictionless, single-sign-on browser access while ensuring user credentials are never persistently stored within the computational environments. \\
        \hline
        \textbf{Centralized Control Hub} & Layer 3: Common Software & Functions as an interactive launchpad and customizable dashboard, allowing users to select specific computing environments upon login. & Offers a familiar, user-friendly interface to configure and launch specialized domain tools directly from the web. \\
        \hline
        \hline
    \end{tabular}
\end{table*}

\subsection{Standardized Tool Connectivity via FastMCP}
To facilitate the seamless integration of diverse independent tools, Sci-Orchestra incorporates FastMCP to handle service connectivity. FastMCP provides a highly efficient, standardized protocol for bridging various data sources, computational scripts, and artificial intelligence models. Rather than developing custom API wrappers for every new scientific service added to the platform, developers can utilize FastMCP to expose their applications uniformly. This dynamic discovery mechanism allows the orchestration engine to instantly recognize the inputs, outputs, and capabilities of newly deployed services. Furthermore, because FastMCP is natively designed to feed context to large language models and automated agents, it effortlessly bridges the gap between traditional scientific computing tools and modern AI-driven workflows. This ensures that data can be cleanly routed between disparate tools within the ecosystem without requiring complex, hard-coded wrappers.

\begin{figure}[t]
    \centering
    \includegraphics[width=1\linewidth]{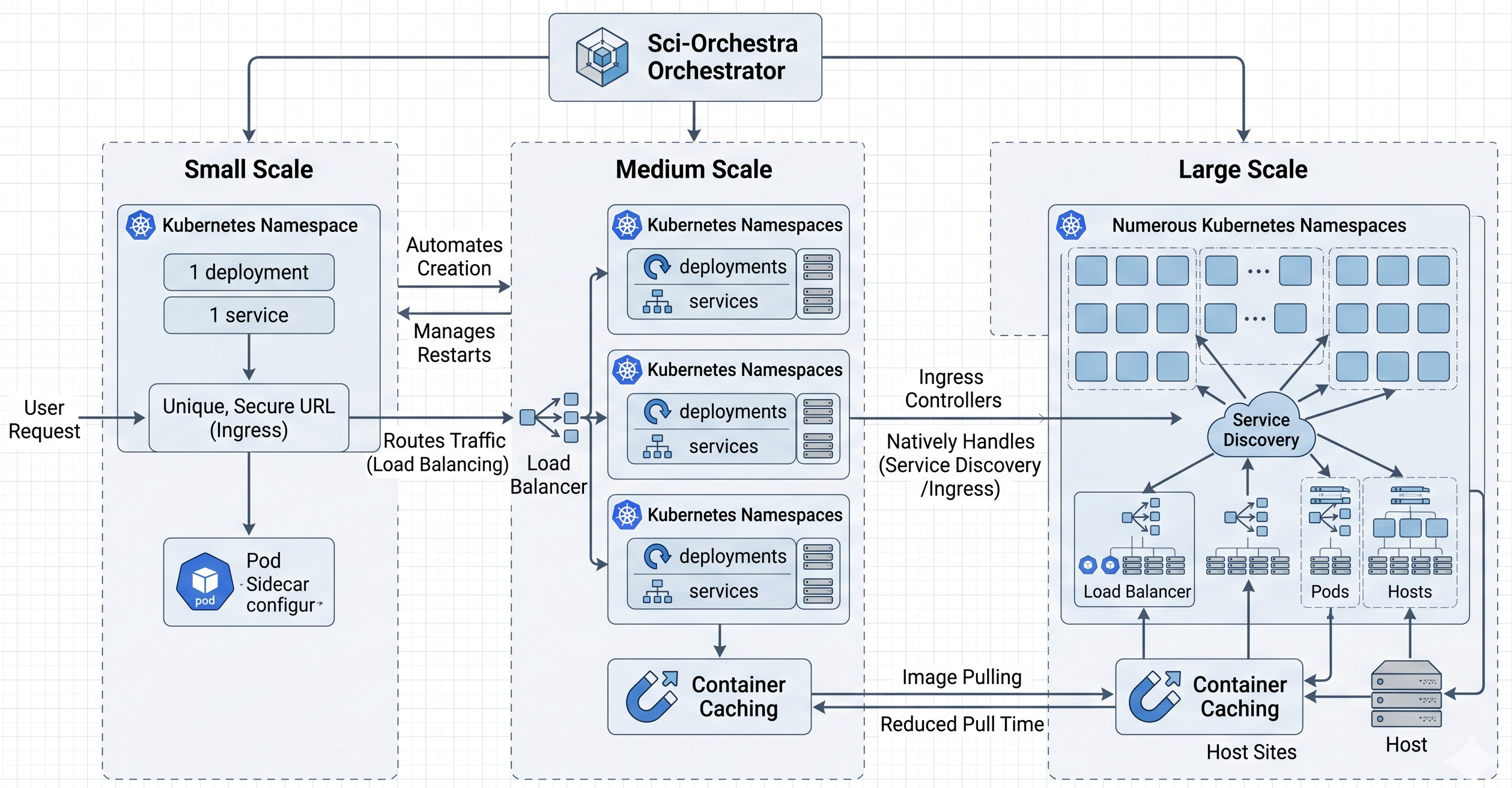}
    \caption{Sci-Orchestra Kubernetes Orchestration and Service Delivery Scalability:  architectural schematic illustrates the multi-tier scalability of the Sci-Orchestra framework, transitioning from individual experimental pods to high-throughput industrial deployments.}
    \label{fig:kubernetes}
\end{figure}

\subsection{Authentication and Application Integration}
Security and seamless access are maintained through an integrated authentication and tool management system. Sci-Orchestra handles user verification using established identity providers, supporting both Google credentials and centralized institutional logins. Furthermore, the framework ensures secure access to event-specific URLs without persistently storing user credentials within the individual computational environments. Once authenticated, users interact with a customizable dashboard that acts as a launchpad for various independent tools. Because Sci-Orchestra is application-agnostic and utilizes standardized connectivity, it can host and manage a wide array of services. Users can deploy standard data science environments, integrate large language models, or launch highly customized, domain-specific services such as autonomous data acquisition and uncertainty quantification with gpCAM~\citep{noack2023exactkernels,noack:2021}, supervised semantic segmentation of thin structures with Rhizonet~\citep{rhizonet}, and zero-shot segmentation with Zenesis~\citep{zenesis2025,zenesis:arxiv:2025}.

\subsection{Kubernetes Orchestration and Service Delivery}
To navigate the modern deployment ecosystem, it is critical to distinguish between container engines and container orchestrators\cite{casalicchio2020state}. While engines like Docker and Podman are responsible for building and running individual isolated software environments—with Podman offering a daemonless, rootless architecture that is critical for HPC security\cite{azab2021containers}—Kubernetes acts as the overarching orchestration layer that schedules, scales, and networks these individual containers across distributed computing clusters.

The backend delivery of our customizable tools relies on advanced Kubernetes service (Fig.~\ref{fig:kubernetes}) management and load balancing handled entirely by Sci-Orchestra. The orchestrator automates the creation of namespaces, deployments, and services, offering users dynamic environments without requiring command-line interaction. It is capable of handling complex application architectures; for example, if a hosted service requires supplementary monitoring tools, Sci-Orchestra can deploy sidecar container configurations, allowing multiple components to run within separate pods under the same container image. Kubernetes natively handles service discovery and ingress configuration, allowing Sci-Orchestra to provide unique, secure URL access for every experimental instance. The system automatically manages and restarts these hosted services, routing traffic through load balancers to eliminate the need for manual daemon configurations. To optimize this deployment process, the infrastructure integrates container caching mechanisms directly at the host sites, significantly reducing the time it takes to pull Docker images.

\subsection{Service Instance Management and Lifecycle Support}
By abstracting the infrastructure layer, Sci-Orchestra allows developers of individual scientific services to iteratively improve their tools without worrying about deployment mechanics. For example, the deployment of a service such as Zenesis can evolve seamlessly through multiple versions on the platform. Deploying such applications as web services allows users to access and utilize them immediately, enjoying automatic updates without needing to manage local installations. Furthermore, the system easily supports concurrent hosting of multiple versions, such as a stable production release and an experimental development version, while allowing administrators to configure specific limits on service usage. To accommodate collaborative environments where multiple researchers might deploy identical tools simultaneously, Sci-Orchestra utilizes descriptive naming conventions and automated URL conflict resolution. This modular approach ensures that the orchestrator maintains strict isolation between instances and distinct boundaries between environments.

\section{Experiments}


To validate the hardware-agnostic capabilities and automated deployment efficiency of the Sci-Orchestra framework, we conducted a comparative deployment experiment using custom scientific software services. We selected two services/tools: a JupyterHub with a workable Notebook and Zenesis-web, a containerized computer vision application that enables image segmentation using zero-shot models and prompting.

Because the Sci-Orchestra architecture isolates the application logic from the underlying infrastructure, the service is inherently hardware-agnostic. To demonstrate this, the services/tools were deployed and tested across two different environments: a generic, independent server and a specialized DOE HPC environment (NERSC Spin and Perlmutter). The experiment tracked the system's ability to interpret the API deployment request, provision the necessary database connections, and expose a secure web endpoint without requiring the developer to modify the underlying application code for either target environment. This demonstrates how a single service template can be generalized to deploy on virtually any backend, seamlessly paving the way for future integrations with commercial cloud providers.

These initial deployments on the baseline server were successfully executed; while configuring this environment was straightforward due to unrestricted root access, it yielded the foundational architectural insights necessary to successfully adapt Sci-Orchestra for the highly regulated, restricted-privilege environments of NERSC Spin and, more recently, NERSC Perlmutter.

\begin{figure}[b]
\centering
\begin{subfigure}[b]{0.2\textwidth}%
    \centering
    \includegraphics[width=\linewidth]{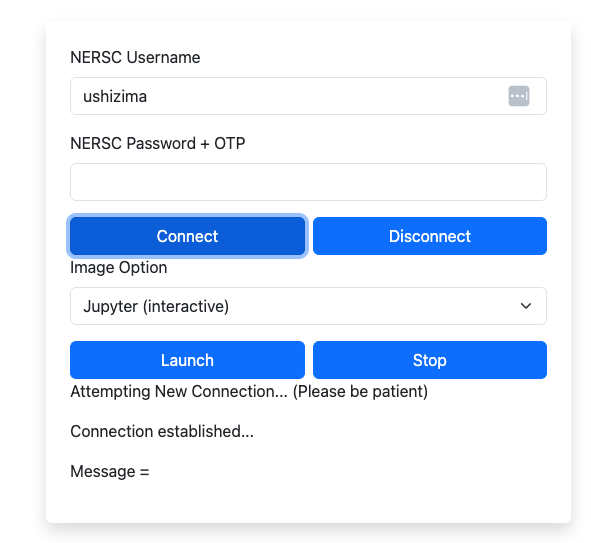}%
    \caption{Logging in.}%
\end{subfigure}\hfill%
\begin{subfigure}[b]{0.4\textwidth}%
    \centering
    \includegraphics[width=\linewidth]{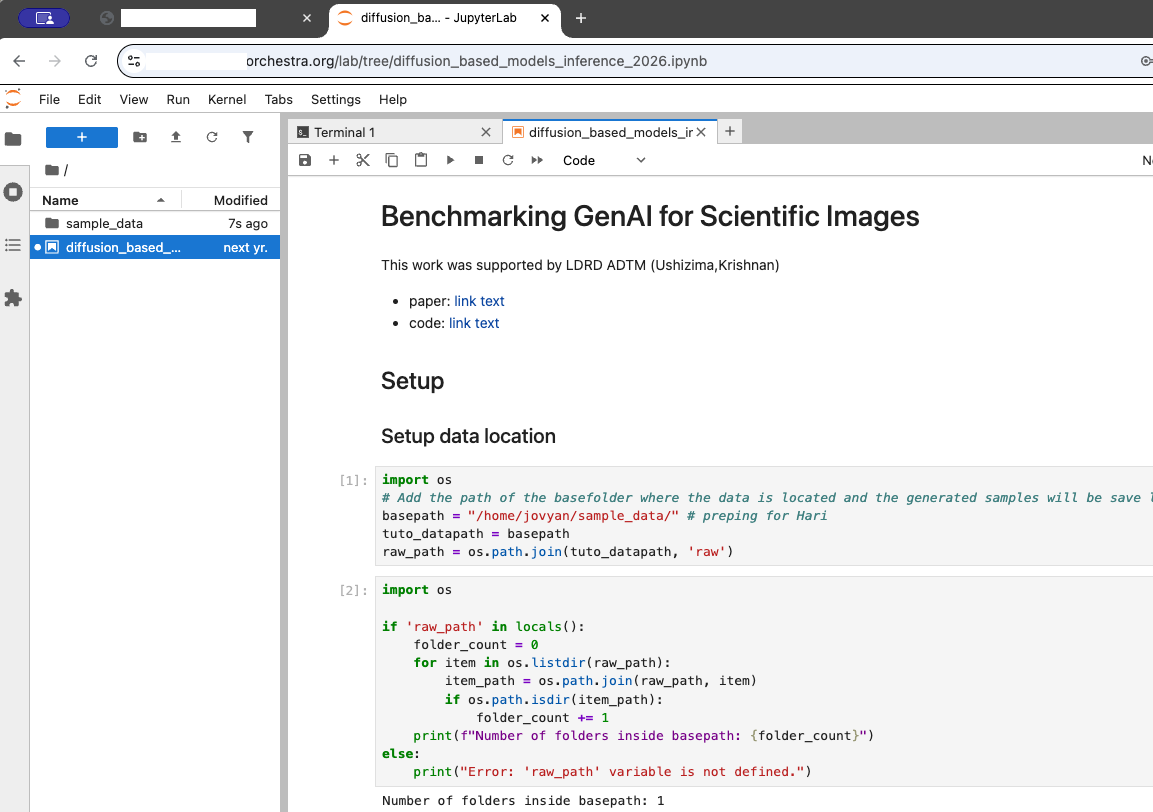}%
    \caption{Running Jupyter Notebook.}%
\end{subfigure}\hfill%
\begin{subfigure}[b]{0.383\textwidth}%
    \centering
    \includegraphics[width=\linewidth]{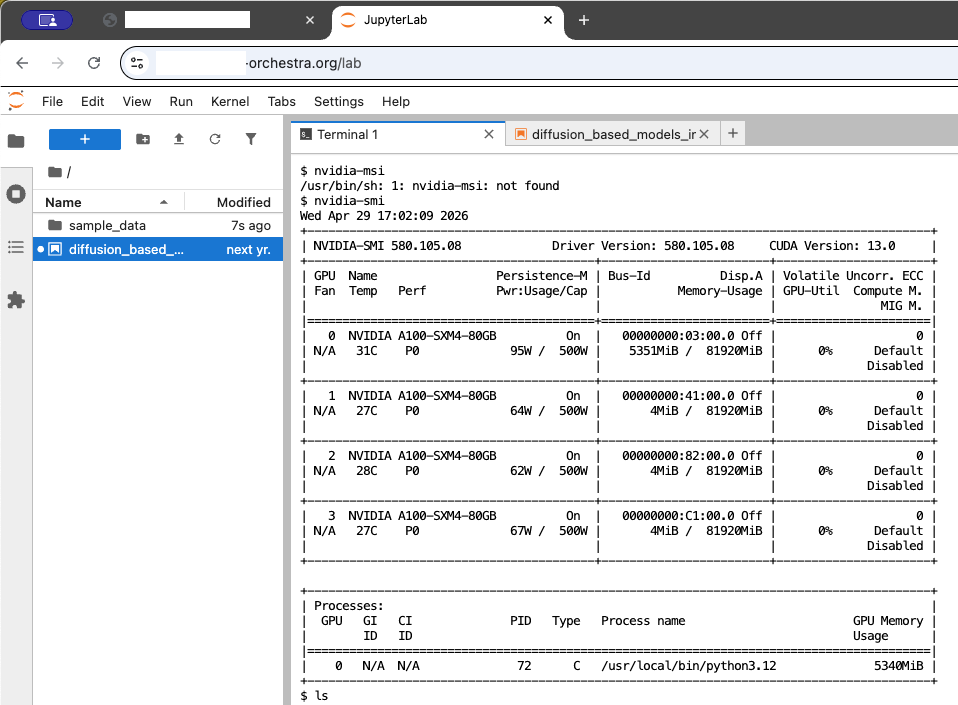}%
    \caption{Computational Resources.}%
\end{subfigure}
\caption{A new and seamless way of  demonstrating software tools in HPC or the cloud: (a) login at NERSC, (b) Jupyter Notebook and respective environment pre-configured, (c) GPU resources allocated automatically.} \label{fig:jupyter}
\end{figure}

\subsection{JupyterHub}
JupyterHub is a multi-user hub that instantiates, orchestrates, and proxies multiple instances of the single-user Jupyter Notebook server. To illustrate the practical efficacy of the Sci-Orchestra framework, Fig.~\ref{fig:jupyter} demonstrates the automated orchestration of a secure, hardware-accelerated Jupyter Notebook environment deployed directly onto HPC infrastructure. On the left, Fig.~\ref{fig:jupyter} (a) captures the streamlined authentication gateway at NERSC, highlighting how the platform abstracts traditional, complex secure shell (SSH) or virtual private network (VPN) protocols into a seamless, web-based federated login experience. 

Upon successful authentication, Fig.~\ref{fig:jupyter} (b) presents the active Jupyter Notebook, showing our benchmark of a series of generative AI schemes applied to scientific images~\citep{jimaging11080252} - the service is instantly provisioned with a fully isolated, pre-configured software environment via backend containerization, completely eliminating the need for the user to manage local dependencies or module loads. Finally, Fig.~\ref{fig:jupyter} (c) displays the underlying computational telemetry, confirming the automatic, policy-compliant allocation of dedicated GPU resources at NERSC Perlmutter to the active session. This end-to-end workflow exemplifies Sci-Orchestra’s core capability: translating a simple user access request into a series of complex orchestration commands that dynamically provision network routing, container execution, and bare-metal hardware acceleration, delivering a production-ready interactive workspace directly to the browser.


Building on this foundational interactive compute capability, the framework easily extends to highly specialized graphical applications; for example, Fig.~\ref{fig:zenesis:tank} depicts the orchestration of Zenesis-web, showcasing the platform's ability to host complex, zero-shot algorithms for single- and multi-instance segmentation workflows in a similarly frictionless manner.

\subsection{Zenesis-web}

\begin{figure}[b]
    \centering
    \includegraphics[width=1\linewidth]{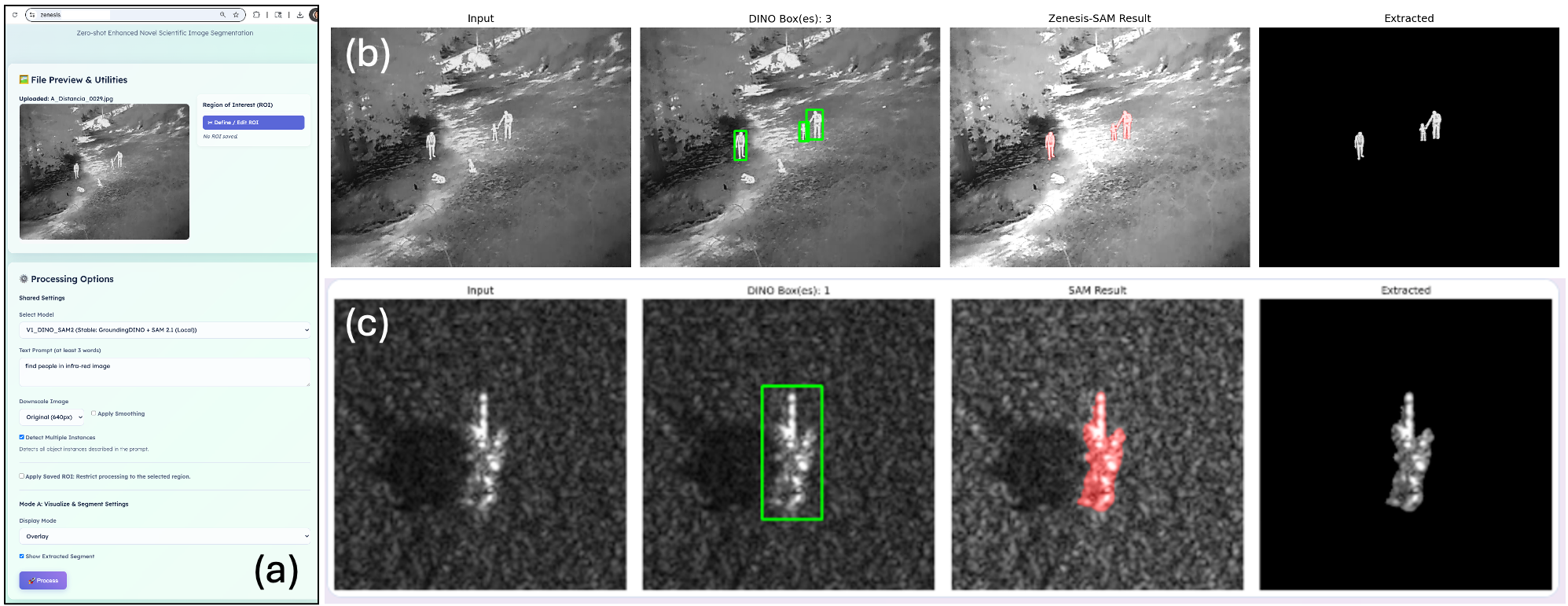}
    \caption{Automated detection using Zenesis-web for zero-shot semantic and object segmentation executed via Sci-Orchestra: (a) Zenesis-web interface; (b) identification of occluded individuals in a forested environment from infrared imagery; (c) military vehicle identification from synthetic aperture radar imagery.}
    \label{fig:zenesis:tank}
\end{figure}

Zenesis-web (Zero-shot Enhanced Novel Scientific Image Segmentation) deploys our Zenesis zero-shot segmentation tool as a web-accessible application. It utilizes multimodal foundation models to segment raw scientific 2D images (Fig.~\ref{fig:zenesis:tank}) and multi-slice volumes, also including natural language text prompts. The application architecture transitioned from localized desktop development to a containerized deployment managed by the Sci-Orchestra platform, allowing it to leverage HPC backend resources dynamically.

As a computationally intensive image analysis platform, Zenesis processes high-resolution datasets with varied bit depths (up to 32-bit). It integrates large-parameter generative models, such as the Segment Anything Model (SAM), to execute multi-stage processing, volumetric segmentation, 3D rendering, and standardized benchmarking (e.g., RefCOCO). The application also incorporates experimental support for in-app parameter-efficient fine-tuning (PEFT) using low-rank adaptation (LoRA)~\citep{hu2022lowrank}. These operations require high-throughput parallelism, volume caching, and persistent storage for model checkpoints—capabilities natively provided by the Sci-Orchestra infrastructure. Executing these tasks on local hardware introduces memory bottlenecks, while traditional HPC job schedulers restrict the continuous execution required for these processes.

Figure~\ref{fig:zenesis:tank}(b) demonstrates the application of Zenesis-web for person segmentation using a thermal infrared (IR) dataset~\citep{infra-red,Ramirez2023} designed for aerial search and rescue operations. Person detection from aerial platforms in wooded areas involves high occlusion variables and small target dimensions relative to the sensor's field of view. While standard RGB datasets for person detection focus on morphological characteristics (size, shape, height), they often fail to account for high-occlusion environments. The utilized IR dataset, collected via an altitude-controlled quadcopter, isolates the thermal signature of individuals partially obscured by vegetation or terrain. Originally developed to train convolutional neural network (CNN)-based models for real-time detection under occlusion, this dataset provides an operationally relevant baseline for evaluating the zero-shot segmentation capabilities hosted via Sci-Orchestra.

Figure~\ref{fig:zenesis:tank}(c) illustrates the segmentation of vehicles from synthetic aperture radar (SAR) imagery sourced from the Moving and Stationary Target Acquisition and Recognition (MSTAR) dataset~\citep{keydel1996mstar,blasch2020mstar,mstar1997}. Developed jointly by DARPA and the Air Force Research Laboratory (AFRL), MSTAR provides a benchmark collection of high-resolution SAR images of military ground vehicles (e.g., tanks, armored personnel carriers, self-propelled artillery). These images were captured at multiple depression angles and azimuth orientations using an X-band SAR sensor with a spatial resolution of approximately 0.3 meters. Providing a standardized set of targets under controlled conditions, MSTAR functions as a primary testbed for evaluating automatic target recognition (ATR) algorithms. Deploying Zenesis-web to process this dataset demonstrates the platform's capacity to handle specialized, non-optical sensor data formats within an interactive web environment.

\section{Discussion}

The deployment of services, such as custom tools in a Jupyter Notebook or Zenesis-web, validates the core architectural premise of Sci-Orchestra: bridging the divide between cloud-native dynamic resource allocation and rigid HPC infrastructure. By abstracting the underlying deployment mechanics, the framework ensures researchers can engage directly with complex visual understanding tasks without executing manual environment configurations. A primary feature enabling this accessibility is the framework's automated ingress routing. While initial validation phases utilize static URLs for controlled testing, the production deployment model automatically generates dynamic, event-specific URLs. This dynamic routing strategy guarantees strict network isolation between multiple concurrent instances, securely mapping individual, browser-based user sessions to their precisely allocated compute and storage resources on the HPC backend. 

Essentially, the architectural abstraction that enables human-in-the-loop interactivity also establishes the programmatic foundation required for automated experimentation. Because the entire deployment lifecycle is governed by an API-first model, the exact mechanisms used to provision a tool dashboard can be triggered by machine learning models, large language models, and autonomous software agents. Through integration with protocols such as FastMCP, autonomous systems can dynamically request HPC resources, orchestrate analytical pipelines, and iterate on data streams without traditional human-in-the-loop bottlenecks. 

This programmatic orchestration is directly applicable to highly dynamic, time-sensitive applications, such as real-time experimental steering at scientific user facilities. In these domains, rapid computational turnaround and strict data security are paramount. The dynamic routing and namespace isolation validated by the Zenesis-web deployment ensure that Sci-Orchestra can instantly provision secure, reproducible analysis environments. Consequently, distributed research teams and autonomous agents can rapidly model physical phenomena or process complex multimodal datasets within a tightly controlled, fully isolated computational ecosystem, significantly accelerating the operational response time of modern automated laboratories.

\section{Conclusion}

Traditional HPC facilities are primarily architected for data-intensive, command-based batch processing executed asynchronously in the background. In contrast, modern scientific workflows—ranging from visual understanding platforms to autonomous AI-driven analytics—require continuous, interactive execution, a paradigm misaligned with conventional multi-node HPC queuing systems. While standard schedulers readily accommodate long-duration background computation, interactive segmentation and real-time data analysis require high-performance compute cycles coupled directly with persistent web accessibility. 

Sci-Orchestra bridges this architectural gap by abstracting the rigid resource allocation protocols of standard HPC environments into a modular, API-first framework. Through advanced containerization, service discovery, and automated network routing, the platform transforms static computational infrastructure into a dynamic, programmable fabric. It provides the persistent network ingress and compute access required for real-time interactive analysis, allowing researchers to manually guide workflows via user-interactive 3D rendering and region-of-interest selection. 

More broadly, by decoupling the scientific software stack from low-level facility toolchains, Sci-Orchestra establishes a comprehensive Science-Platform-as-a-Service. This infrastructure not only lowers the barrier to entry for deploying complex, interactive web applications on HPC systems, but also provides the programmatic backbone necessary for the integration of autonomous agents. By enabling both human researchers and automated systems to instantly provision secure, multi-tier workflows, Sci-Orchestra equips the scientific community with the agile computational ecosystem required to address increasingly complex and time-critical research challenges.

\section{Future Developments and Applications}
While the current iteration of Sci-Orchestra successfully automates browser-based deployments on systems such as NERSC Spin and Perlmutter, the framework's decoupled architecture allows for significant future expansion. A primary focus for near-term development is the expansion of multi-system orchestration capabilities. We intend to register additional external platforms—such as Amazon Web Services (AWS), the ALCF, and the broader NRP—as dedicated sub-orchestrators. This evolution will allow the main Sci-Orchestra engine to intelligently route user requests to highly specialized hardware, such as GPU-backed HPC nodes or memory-optimized cloud instances, based strictly on the schema constraints of the requested event.

Additionally, we plan to improve data management by integrating native S3-compatible storage solutions directly into the orchestration lifecycle, allowing for seamless data persistence, artifact tracking, and retrieval across isolated computing centers. Concurrently, we will expand the platform's automation capabilities by introducing template workflows and deeper support for autonomous AI agents. By allowing agents to leverage FastMCP and the API-first design, Sci-Orchestra will eventually enable AI-driven systems to autonomously compose, deploy, and monitor complex, multi-facility scientific experiments without human intervention. This programmatic capability serves as the foundational infrastructure for the next generation of "smart labs" and self-driving laboratories, where closed-loop systems dynamically adjust experimental parameters, synthesize new data, and route analytical workloads in real-time.

To demonstrate the versatility of Sci-Orchestra across a wide range of scientific domains, it is instructive to examine how distinct workflows currently leverage the platform's core, reusable components. For instance, the automation of distributed federated learning services presents a unique infrastructure challenge. In projects such as Advanced Privacy-Preserving Federated Learning (APPFL)~\citep{appfl_ai}, the primary objective is to isolate access to sensitive datasets by sharing only trained model parameters rather than raw information, enabling secure aggregate findings. Sci-Orchestra directly supports this paradigm through its robust authentication and authorization framework. The platform seamlessly manages the secure launch of isolated computational services, restricts access exclusively to authorized research group members, and provides the orchestration engine necessary to scale these secure enclaves across both private clusters and public cloud deployments.

Similarly, the framework is highly applicable to end-to-end experimental control pipelines, such as those required by the Berkeley Synchrotron Infrared Structural Biology (BSISB) Imaging Program~\citep{hoying:2023,bsisb_imaging}. BSISB workflows span the entire experimental lifecycle, from direct scientific instrument control to complex post-experiment analytics. This necessitates unified, secure access for visualization, hardware control, logging, and data provenance tracking. Sci-Orchestra facilitates these comprehensive workflows by orchestrating off-the-shelf tools—such as Virtual Network Computing (VNC) containers for remote visualization—to control specialized proprietary software such as OMNIC, while simultaneously executing downstream data analytics at computing centers such as NERSC. By providing a centralized, API-driven command and control ecosystem, Sci-Orchestra consolidates real-time instrument interaction and high-performance analytics into a single, cohesive portal for the research team.

\section*{Acknowledgments}
This work was supported by the Laboratory Directed Research \& Development (LDRD) Program Analytics through Diffusion Transformer Models (ADTM). Partial support was also provided by the US Department of Energy (DOE) Office of Science Advanced Scientific Computing Research (ASCR) and the Office of Basic Energy Sciences (BES) under Contract No. DE-AC02-05CH11231 to the Center for Advanced Mathematics for Energy Research Applications (CAMERA); the DOE ASCR funding project Autonomous Solutions for Computational Research with Immersive Browsing \& Exploration (ASCRIBE); and the DOE Office of Biological and Environmental Research (BER)- and ASCR-funded projects Restor-C and Twins. We also utilized HPC resources at the National Energy Research Scientific Computing Center (NERSC) at Berkeley Lab.